\documentclass[conference]{IEEEtran}
\ifCLASSINFOpdf
  \usepackage[pdftex]{graphicx}
\else
\fi
\usepackage{subcaption}
\usepackage{float}

\begin{document}

\IEEEoverridecommandlockouts
\IEEEpubid{\begin{minipage}{\textwidth}\ \\[12pt] 
\\\\\\
  Personal use of this preprint copy is permitted. Republication, redistribution\\
  and other uses require the permission of IEEE.\\
  This paper was published within the proceedings of the 25th European\\ Signal Processing Conference (EUSIPCO), 2017, Kos, Greece\\ 
  DOI: 10.23919/EUSIPCO.2017.8081218, \copyright 2017 IEEE.\\
\end{minipage}} 
%
\title{Qualitative Assessment of Recurrent Human Motion}

\author{\IEEEauthorblockN{Andre Ebert, Michael Till Beck, Andy Mattausch, Lenz Belzner, and Claudia Linnhoff-Popien}
\IEEEauthorblockA{
Mobile and Distributed Systems Group\\
Institute for Computer Science\\
Ludwig-Maximilians-University, Munich, Germany\\
Email: andre.ebert@ifi.lmu.de, michael.beck@ifi.lmu.de, andy.mattausch@ifi.lmu.de, belzner@ifi.lmu.de, linnhoff@ifi.lmu.de}
}


%


\maketitle



\begin{abstract}
Smartphone applications designed to track human motion in combination with wearable sensors, e.g., during physical exercising, raised huge attention recently.
Commonly, they provide quantitative services, such as personalized training instructions or the counting of distances.
But qualitative monitoring and assessment is still missing, e.g., to detect malpositions, to prevent injuries, or to optimize training success.   
\par
We address this issue by presenting a concept for qualitative as well as generic assessment of recurrent human motion by processing multi-dimensional, continuous time series tracked with motion sensors.
Therefore, our segmentation procedure extracts individual events of specific length and we propose expressive features to accomplish a qualitative motion assessment by supervised classification.
We verified our approach within a comprehensive study encompassing 27 athletes undertaking different body weight exercises.
We are able to recognize six different exercise types with a success rate of 100\% and to assess them qualitatively with an average success rate of 99.3\%.
\end{abstract}



%
\IEEEpeerreviewmaketitle


\section{Introduction}\label{sec:introduction}
Regular physical exercising improves an athlete's health and well-being; sufferings from chronic diseases or even the Alzheimer's disease are lowered \cite{radak2010exercise}.
In that context, mobile phone applications for training support (e.g., running, CrossFit, etc.) became popular.
They provide customized workout plans, detailed exercise instructions as well as quantitative and statistical functions.
But by providing know-how about challenging exercises without supervision to non-experienced athletes arises new problems.
Wrong execution of exercises, malpositions, or the absence of sufficient warming up phases may lead to less training success or even to serious injuries. 
Especially non-experienced athletes are likely to harm themselves during an unsupevised workout \cite{koplan1985risks}.
\par
\begin{figure}[!t]
\centering
\includegraphics[width=3.4in]{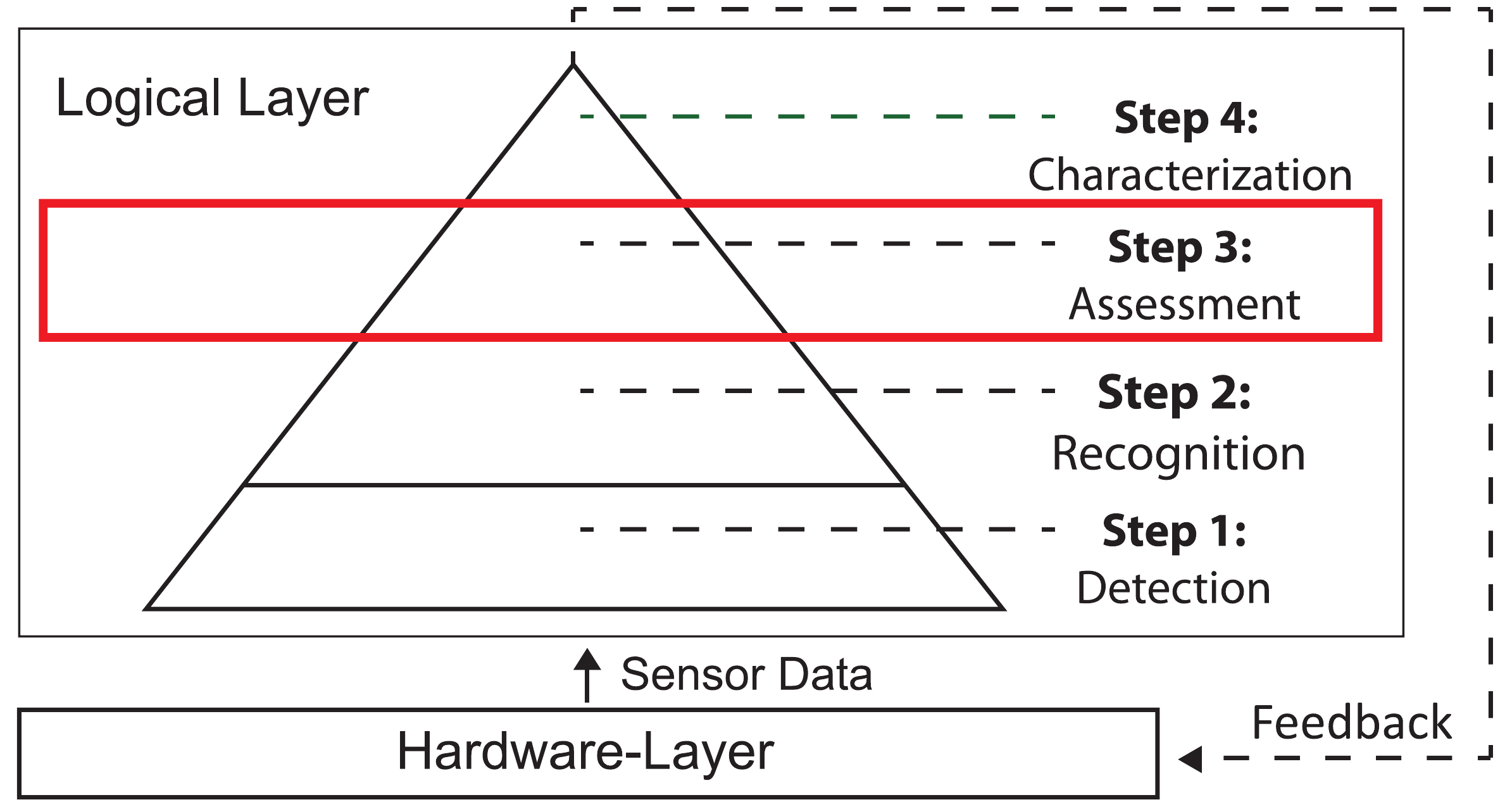}
\caption{Four fundamental steps of human motion analysis within the logical layer, the hardware layer functions as a sensor and feedback provider, as proposed by \cite{ebert2017sensx}.}
\label{fig:fundamentalsteps}
\end{figure}

We believe that a pro-active and automated monitoring reduces such injuries drastically while a training's success could be improved significantly. Moreover, a generic concept capable of recognizing and assessing various recurrent human motions is also applicable in other areas, e.g., medical observations, gait analysis or optimization of workflows.  
To address this unsolved issue we previously introduced SensX, which is a distributed sensor system for capturing and processing human motion \cite{ebert2017sensx}.
We established a paradigm for qualitative analysis of human motion consisting of four fundamental steps (see Figure \ref{fig:fundamentalsteps}): (1) \textit{Detection} of a motion event, (2) its \textit{Recognition}, (3) its qualitative \textit{Assessment}, and (4) the \textit{Characterization} of reasons for a specific assessment. Step (1) and (2) were treated within \cite{ebert2017sensx}, while this paper focuses on step (3) by using the existing SensX architecture as a basis.
\setlength{\textfloatsep}{5pt}
\par
Thereby, our contributions within this paper are as follows: 
\begin{itemize}
\item We propose a novel concept for qualitative assessment of complex, recurrent human motion.
\item It covers the extraction of multi-dimensional motion events into segments of individual length.
\item An expressive feature set as well as a system for supervised classification are selected and implemented.
\item To validate our concept, we conducted an comprehensive exemplary study with 27 athletes executing more than 7,500 repetitions of six different types of body weight exercises.
\item We present state-of-the-art results concerning the assessment of human motion as well as for human activity recognition on basis of motion sensor data.
\end{itemize}

\section{Related Work}\label{sec:relatedwork}
In the following, we provide a brief overview on related work concerning 1) segmentation, 2) recognition and 3) assessment of human motion. 
Thereby, we are focusing on complex motion sequences which are described by multiple, coordinated movements conducted by several extremities at the same time (e.g., body weight exercises), instead of more simple activities which have often been subject to activity recognition within previous research (e.g., walking or sitting).
\par
1) Before analysis and assessment of reoccurring events within multi-dimensional time series become feasible, they need to be extracted into segments first.
Bulling et al. name (1) sliding windows, (2) energy-based segmentation, (3) rest-position based segmentation, and (4) the use of additional sensors or context resources as applicable procedures \cite{bulling2014tutorial}. 
Sliding window approaches (1) move a window of static size sequentially across an incoming stream of data and extract the window's current content for further analysis.
E.g., authors of RecoFit and ClimbAX used a 5s sliding window which they moved in discrete steps across a motion data stream \cite{morris2014recofit, ladha2013climbax, ding2015femo}.
These approaches offer valuable ideas for our segmentation concept. 
Still, due to the absence of a length adjustment to an events actual duration, they do not cover all of our needs.
The actual start and endpoints of short events (e.g., a pushup) are not captured accurately, which leads to noise within an event's segment (e.g., fragments of preceding or following events).
This noise may disturb the qualitative assessment process of a specific event significantly. 
Energy-based solutions (2) perform well for segmentation of long term activities, which are describable by different energy potentials (e.g., sitting, running).
We examine individual repetitions of short movements - their energy potential is not diverse enough from each other and is not suitable for segmentation.  
(3) Rest-position based segmentation is also not feasible, for there are no rest positions within a continuous event set. 
The use of external information sources (4), e.g. GPS, is not suitable for such fine-grained movements targeted by us.
Other approaches also facilitated manual segmentation, which is not suitable for great numbers of events or realtime analysis \cite{mortazavi2014determining}.
The review of these procedures led us to the necessity of developing an individualized segmentation process, which considers our requirements concerning fine-grained, dynamic and accurate extraction of complex and multi-dimensional motion events.  
\par
2) Quantitative counting of repetitive activities as well as its recognition and classification are well-trodden fields of research, which is why we do not present much work bound to that topic within this paper. 
E.g., Jiang et al. recognize simple activities like lying, walking, and sitting, while Morris et al. are dealing with more complex exercises \cite{morris2014recofit, jiang2015human}. 
Facilitated techniques are neuronal networks as well as typical classifiers for supervised learning and combinations of both.
\par
3) In contrast to that, qualitative assessment of human motion data was examined only sparsely, yet. 
Ladha et al. as well as Pansiot et al. assessed the performance of climbers by extracting and analyzing features such as power, control, stability, and speed without examining individual climbing moves \cite{ladha2013climbax, pansiot2008climbsn}. 
Their work provides valuable information concerning the handling and preprocessing of motion data. Still, it does not allow the assessment of individual movements of specific extremities within a chain of multiple climbing features. 
Velloso et al. assessed repetitions of recurrent motion by recording five wrongly executed weight-lifting exercises and classified them afterwards by template comparison \cite{velloso2013qualitative}.
Though they were able to classify exercise mistakes with a success rate of 78\%, their template-based approach is only able to identify a fixed number of predefined error cases.
Thus, it is not suitable for generic assessment of human motion.
GymSkill is a system for qualitative evaluation of exercises conducted on a balance board \cite{moller2012gymskill}. 
Exercises are examined and assessed with an individualized Principal Component Analysis, but GymSkill is bound to analysis in combination with a balance board. 
Therefore, it is also not capable of generic motion analysis.
\par
Concluding, we are not aware of a concept which enables qualitative assessment of human motion in a generic way and without being bound to predefined motions or equipment.
As a solution, we now present a novel approach for tracking, recognizing and assessing human motion. 

\section{An approach for qualitative assessment of human motion}\label{sec:concept}
In the following, we explain our advance for extracting recurring events out of multi-dimensional time series. Afterwards, we describe the preprocessing and selection of expressive features to prepare a qualitative assessment via supervised learning. 
Data input of our analysis procedure are 30 individual streams of motion data, see \cite{ebert2017sensx} for technical details. Therefore, five sensor devices are tracking acceleration and rotation information in X-, Y-, and Z-dimension. 
Two are fastened on the tracked person's ankles, two are attached to the wrist, and the fifth is worn on the chest in combination with a processing unit. 

\subsection{Preprocessing and segmentation}\label{subsec:segmentation}
To extract individual events out of the incoming multi-dimensional signal set, we developed a dynamic and multi-peak-based segmentation algorithm, which is capable of segmenting heterogeneous sets of motion events individually: each signal in each segment may be of specific length and contains all information about exactly one rotation or acceleration axis of exactly one specific event. 
All extracted event segments are also of individual temporal length in comparison to each other.
\par
\begin{figure}[!t]
\centering
\includegraphics[width=3.4in]{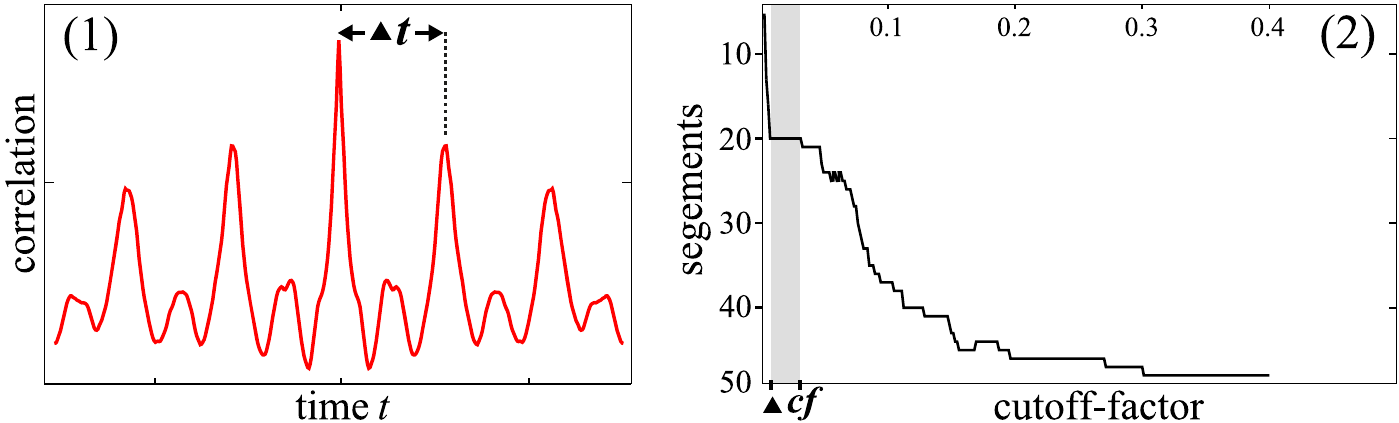}
\caption{Auto-correlated signal of a set of mountain climbers (1) and determination of the ideal cutoff factor \textit{cf} (2).}
\label{fig:cfandwsize}
\end{figure}

To identify individual motion events, we first examine the most meaningful signal vector within a signal set. 
Typically, this signal contains the highest dynamics and variance within its values and allows a distinct identification of a segment's $S_1$ start point $t_{s1}$ and its end point $t_{e1}$. 
For that, we are calculating the standard deviation $\sigma$ of all signals, whereby $X$ defines the current signal, $x_i$ is the $i$-th measured value and $\mu$ is the expectancy value:
\vspace{-4mm}
	$$\sigma=\sqrt{Var(X)} = \frac{\sum\limits_{i=1}^n (x_i - \mu)^2}{n}$$
The signal with the highest deviation is taken for further analysis.
We assume that every type of motion event can be described by an individual set of local extrema and we use these sets to identify distinct events.
\begin{figure}[!t]
\centering
\includegraphics[width=3.4in]{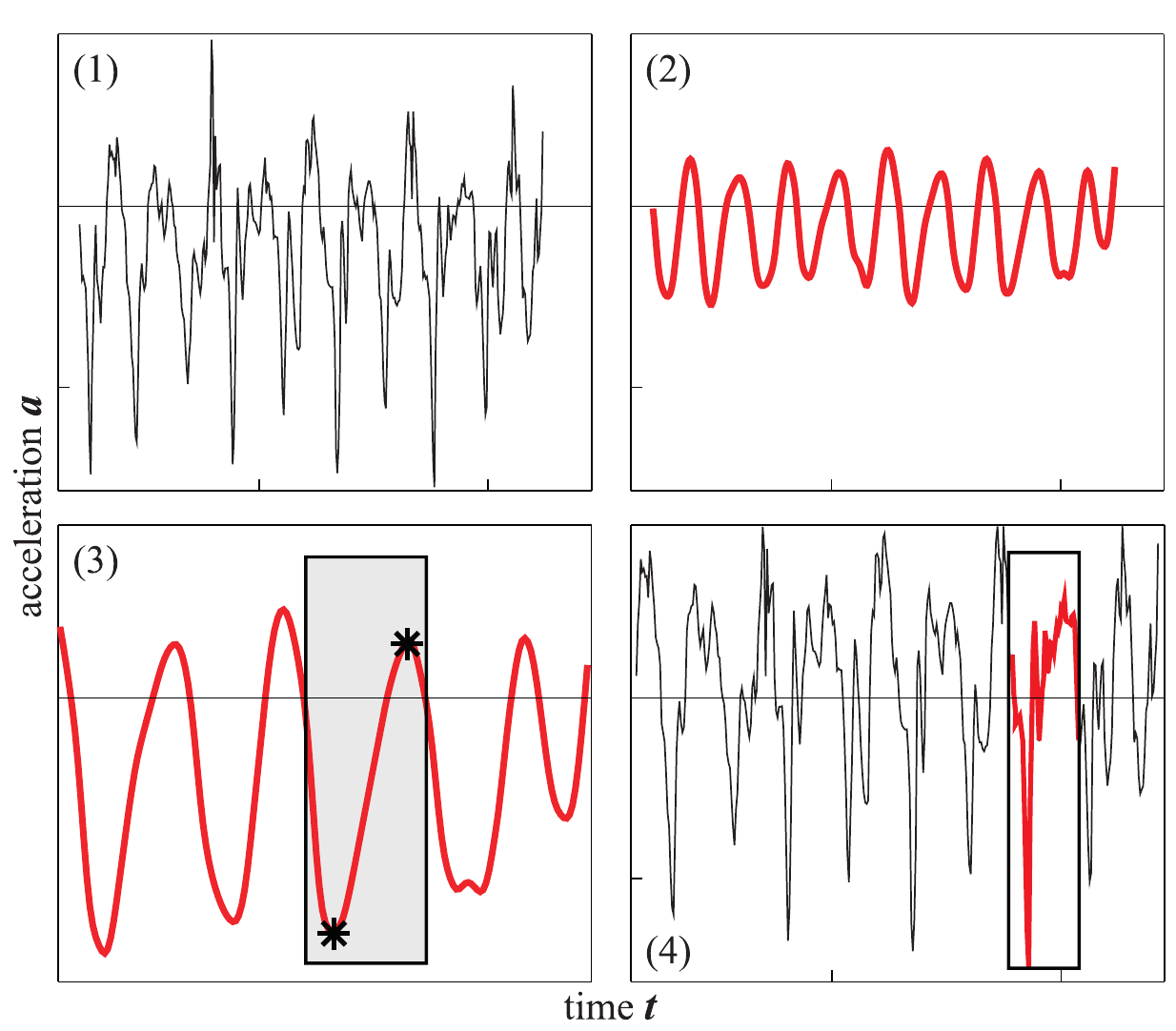}
\caption{Step by step procedure of segmenting recurrent motion events of individual length out of a set of bicycle crunches.}
\label{fig:segmentationprocess}
\end{figure}
\setlength{\textfloatsep}{3pt}
Figure \ref{fig:segmentationprocess} depicts the whole segmentation process for a set of bicycle crunches. Within this setting, the acceleration of the ankle sensors along the X-axis proved to provide the most meaningful signal.
As depicted in Figure \ref{fig:segmentationprocess}-1, signals of individual repetitions contain a high amount of noise as well as unique peaks which are not representative for a specific class of movements. 
These peaks may contain information which is critical for assessing a movement in terms of quality, but they are irrelevant for segmentation. 
That is why we designed and applied an aggressive Butterworth low pass filter to the signal (see Figure \ref{fig:segmentationprocess}-2). 
Thus, all information unnecessary for segmentation is extinguished and only essential periodicities are left.
The filter's cutoff-frequency $f_c$ is determined by multiplying the sampling frequency $f_s$ with a cutoff factor $f_c = f_s * cf$, which is essential for the filter's effect onto the signal.
Figure \ref{fig:cfandwsize}-2 shows the results of the empirical determination of $\Delta cf$.
It indicates that our sensor setup $cf$ must be within a range of 0.0065-0.025 in order to recognize all individual event occurrences within a set of 20 repetitions. Thereby, a different $cf$ setup is used for each individual exercise.
\setlength{\textfloatsep}{5pt}
Due to the low pass filtering in combination with the usage of extrema patterns, the identification of the individual duration $\Delta t$ of a segment as well as its starting point $t_{sx}$ and its ending point $t_{ex}$ (see Figure \ref{fig:segmentationprocess}-3) becomes feasible. 
This is achieved as follows: e.g., a bicycle twist can be described by a set of one local minimum and one local maximum. 
Other movements may be characterized by differing combinations of multiple local extrema, as depicted by the auto-correlation for a set of mountain climbers in Figure \ref{fig:cfandwsize}-1 (at least two different signal parts are identifiable for this example).
Our filtered signal is now scanned sequentially for this pattern.
When a new occurrence is detected, a window of the size of the estimated event length is applied to the signal in a first phase.
The estimated event length $\Delta t$ is derived from the event sets auto-correlation, as depicted in Figure  \ref{fig:cfandwsize}-1.
But since every repetition is of individual length and content, we need to adjust the segments start and end points individually within a second phase. 
Precondition for the following is the assumption, that origin and terminus of a repeated motion event is located at the signal's zero crossing in between the rest periods.
Now we check if the segment window encompasses the demanded number and types of extrema. 
If not, we sequentially add sub segments of a predefined length $l$, until the relevant extrema pattern is matched. 
After this matching phase, some fine tuning is undertaken to capture the exact segment ending: if the last element within the segment is a positive value, we wind forward and add single samples until we reach the next zero crossing. 
Otherwise, if the last element within the segment is a negative value, we wind back to the last zero crossing and remove all values on this way. 
After determining the individual length of the current segment, we only keep the timestamps of its starting and its endpoint $t_{sx}$ and $t_{ex}$. 
Subsequently these are used to cut out the specific segment from the slightly smoothed original signals (see Figure \ref{fig:segmentationprocess}-4), which still contain all important movement information.
Output of this procedure is a quantity $S = \{{S_1,S_2,S_3,S_x\}}$ of event segments of differing length, whereby each segment has its exact borders and contains only information of exactly one motion event.

\subsection{Feature selection and labeling}\label{subsec:featureselection}
\begin{figure}[!t]
\centering
\includegraphics[width=3.4in]{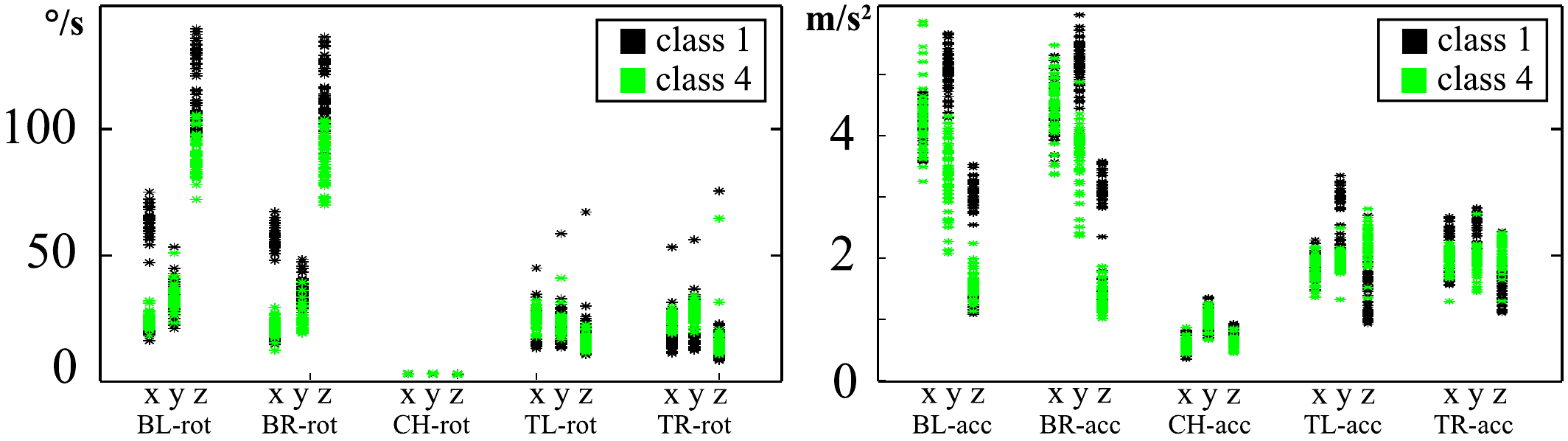}
\caption{Standard deviations of all rotation and acceleration signals of a set of 200 lunges labeled class 1 and class 4.}
\label{fig:lungesdeviation}
\end{figure}

\begin{figure}
\begin{subfigure}{.24\textwidth}
  \centering
  \includegraphics[width=0.9\linewidth]{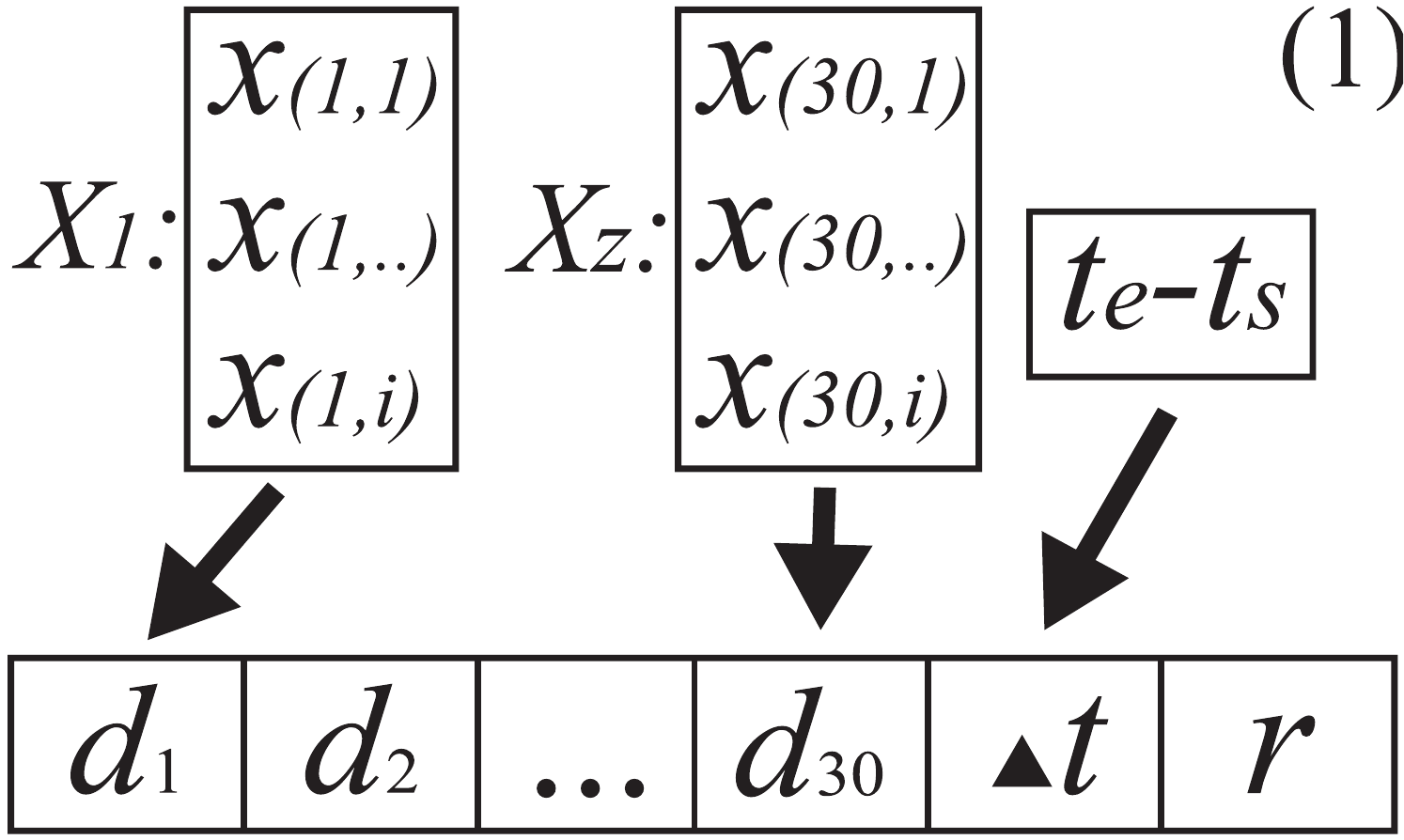}
\end{subfigure}%
\begin{subfigure}{.24\textwidth}
  \centering
  \includegraphics[width=.8\linewidth]{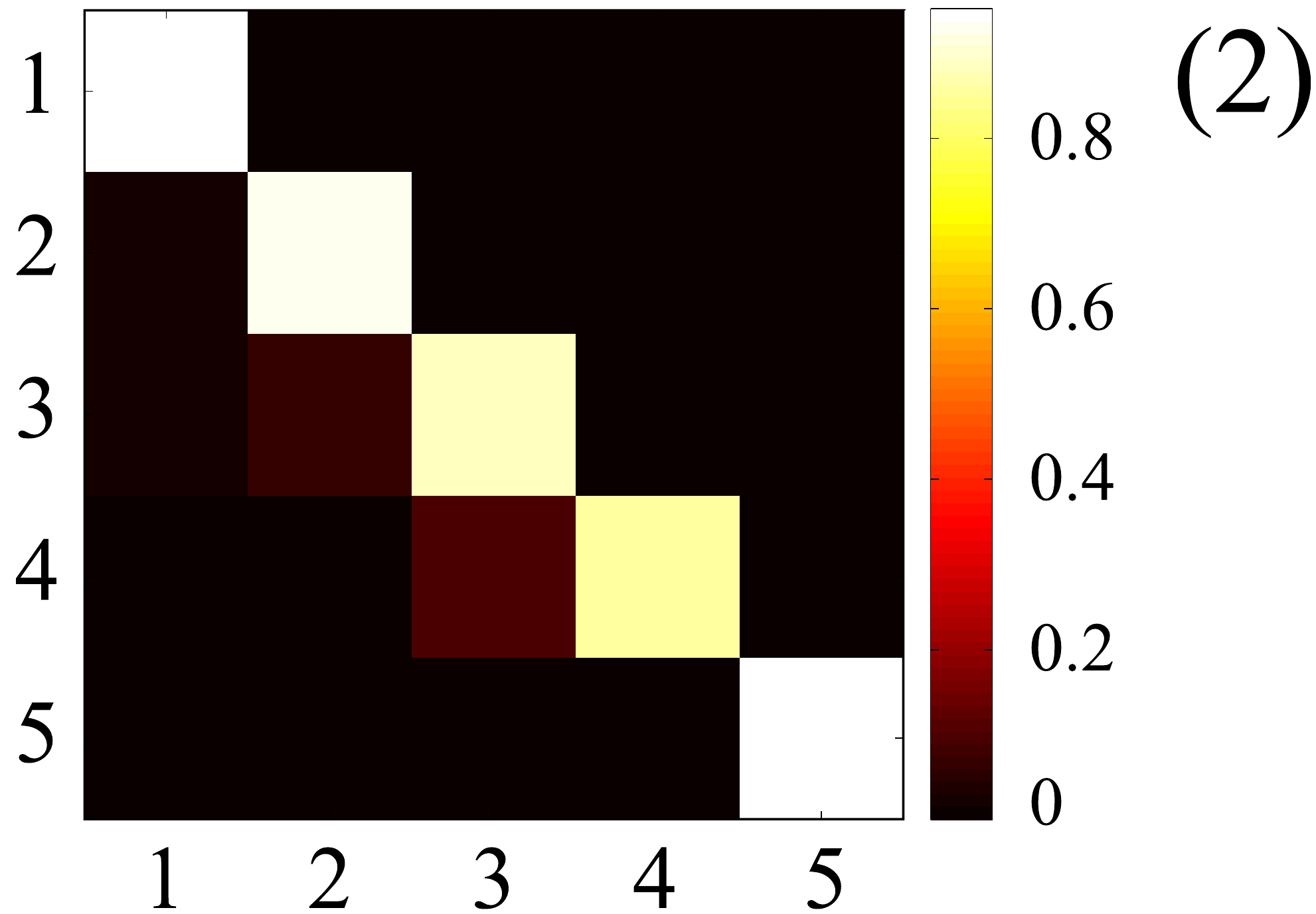}
\end{subfigure}
  \caption{Components of the feature vector, which is describing one individual motion event (1) and the confusion matrix visualizing the results for qualitative assessment of bicycle crunches (2).}
\label{fig:featuresandlunges}
\end{figure}

Commonly, a feature vector within machine learning scenarios consists of a fixed number of features describing one instance.
Due to the fact that all of our activity segments are of individual length and consist of 30 individual signals, this issue is challenging. 
If we use the segmented time series directly for feature set creation, their length would need to be trimmed or interpolated to match the fixed length of a feature vector. 
Interpolation would result in unwanted artificial noise, trimming could lead to the loss of important information and finally, all preceding efforts to extract each event into a segment of individual length would be worthless. 
Furthermore, one event of the dataset that we recorded for evaluation (see Section \ref{sec:evaluation}) consists of 3455 sample values in average (roughly 155 per signal). 
Building feature vectors of this length and greater leads to massive computational load during classification. 
To overcome these challenges were exploited some observations we made during the examination of our dataset: Figure \ref{fig:lungesdeviation} visualizes the standard deviations of the acceleration and rotation signals of 100 randomly selected lunges labeled with quality class 1 (very good) and 100 lunges labeled with quality class 4 (poor).
In general, lunges labeled class 1 show a much higher deviation in rotation and acceleration values for the users feet (bottom-left (BL), and bottom-right (BR)), while class 4 values are significantly lower. 
This is because a proper lunge is described by a big step forward as well as bringing one knee nearly to the ground, which results in a greater movement energy while not decently conducted lunges result in less energy within these signals. 
The same happens forwards (Y-Axis) and downwards (X-axis) for the wrist's acceleration (top-left (TL) and top-right (TR)), which are placed onto the users hip during the workout. 
In contrast to that, the rotation is low for proper lunges and higher for the improper ones. 
This is related to a smooth movement, conducted by skilled athletes and more unsteady movements conducted by unskilled athletes. 
Due to a relatively smooth and steady movement of the athletes torso, the chest sensor (CH) did not provide significant information concerning this activity. 
These observations show that even the individual signal's standard deviation contains enough information to assess an activity in a qualitative way. Based on this cognition we designed a feature vector to describe each individual activity instance. 
It contains the standard deviations of all 30 signals plus the time interval $\Delta t$ of the specific instance in milliseconds (see Figure \ref{fig:featuresandlunges}-1). All in all, one event out of our evaluation data set (see Section \ref{sec:evaluation}) consists of an average of 3.364 sampling values -- by utilizing the procedure described above, we are able to compress this information by a ratio of 1:109. 
Additionally, we added a label \textit{r} concerning the individual motion events' quality rating from 1-5 (very good to very poor, see Section \ref{subsec:dataset}.  
\setlength{\textfloatsep}{5pt}

\section{Evaluation}\label{sec:evaluation}
In this section, we first describe the setup of our study. 
Subsequently, we present the results of our evaluation and give insights into the performance of the segmentation algorithm. 

\subsection{Dataset}\label{subsec:dataset}
For evaluation, we recorded six body weight exercises (crunches (cr.), lunges (lu.), jumping jack (j.j.), mountain climber (m.c.), bicycle crunches (b.c.), and squats (sq.)) conducted by 27 athletes of male and of female sex and aged from 20 till 53. 
Each athlete had to complete 3 sets with 20 repetitions of each exercise; between the individual sets we scheduled a mandatory break of 30\textit{s}. 
An instruction video was shown to the athletes for each exercise and in prior of its execution. 
All in all we tracked motion data of 7,534 individual exercise repetitions -- additionally, all conducted exercises were taped on video for later on labeling by experts. 
The labeling range is 1 (very good) to 5 (very bad). 
The data is labeled as follows: all exercises were labeled initially with class 1. 
For each mistake (each specific deviation from the video instructions), e.g., steps are too small for a mountain climber, the initial class gets added 0.5 (small deviation) or 1 (severe deviation) error points. The final class is the rounded result of the overall error score. 
Hence, completely different errors during the performance of a motion event may lead to the same error score and therefore the same quality class.

\subsection{Qualitative motion assessment}\label{subsec:assessment}
All in all, we used two different classification approaches for supervised learning, one with 1) manual and one with automated hyper parameter optimization. 
Within 1), we manually configured four popular classification algorithms for human activity recognition (see Section \ref{sec:relatedwork}): the decision tree driven Random Forest (RF) and C4.5, a Support Vector Machine (SVM) classifier, and the Naive Bayes (NB) algorithm.
\begin{table}[htb]
\scriptsize
\centering
\setlength{\tabcolsep}{0.5em} 
{\renewcommand{\arraystretch}{1.2}
\begin{tabular}{|l|l|l|l|l|l|l|l|l|}
   \hline
\textbf{classifier} & \textbf{cr.} & \textbf{lu.} & \textbf{j.j.} & \textbf{b.c.} & \textbf{sq.} & \textbf{m.c.} & \textbf{average} & \textbf{duration} \\ 
	\hline
RF & 88.0\% & 90.0\% & 92.1\% & 92.1\% & 93.4\% & 82.5\% & 89.7\% & 447$ms$ \\
C4.5 & 79.1\% & 80.5\% & 83.6\% & 82.1\% & 84.6\% & 67.9\% & 79.6\% & 65$ms$ \\
SVM & 73.2\% & 80.8\%  & 85.0\% & 85.7\% & 80.2\% & 60.2\% & 77.5\% & 83$ms$ \\
NB & 54.3\% & 70.3\% & 72.5\% & 76.3\% & 58.5\% & 54.6\% & 64.4\% & 83$ms$ \\
\hline
\end{tabular}
}
\vspace{-4mm}
\\\bigskip
\small
\caption{Correct classification rates for qualitative assessment with manual classifier selection and configuration.}
\label{tab:qualres}
\end{table}
\vspace{-4mm}
Table \ref{tab:qualres} presents the performance of manual classifier selection within a 10-fold cross validation. 
RF provides the best results with an average correct classification rate of 89.7\% while taking 447$ms$ for building its evaluation model. 
NB performed way faster, but with worse results.
\begin{table}[htb]
\scriptsize
\centering
\setlength{\tabcolsep}{0.5em} 
{\renewcommand{\arraystretch}{1.2}
\begin{tabular}{|l|l|l|l|l|l|l|l|}
   \hline
 & \textbf{cr.} & \textbf{lu.} & \textbf{j.j.} & \textbf{b.c.} & \textbf{sq.} & \textbf{m.c.} & \textbf{average} \\ 
\hline
Success Rate & 100\% & 100\% & 99.9\% & 96.0\% & 100\% & 100\% & 99.3\% \\
Training time & 2,071$ms$ & 673$ms$ & 1$ms$ & 2$ms$ & 757$ms$ & 1$ms$ & 584$ms$ \\
Classifier & RF & RF & k* & IBk & RF & k* & -\\ 
\hline
\end{tabular}
}
\vspace{-4mm}
\\\bigskip
\small
\caption{Correct classification rates for facilitating a hyper parameter optimization with Auto-WEKA.}
\label{tab:qualresauto}
\end{table}
\vspace{-4mm}
\par
Approach 2) facilitates Auto-WEKA as hyper parameter optimization layer for automated selection of appropriate classifiers and hyper parameter tuning \cite{thornton2013auto}. 
Table \ref{tab:qualresauto} shows significantly improved results by facilitating RF, k-nearest neighbor (IBk), and K-Star (k*), also for a 10-fold cross validation.
Four out of six exercise types are assessed correctly with a success rate of 100\% while the average rate is about 99.3\%. 
Despite varying time spans for different classifiers, all test models except one were trained within less than a second for thousands of event instances. 
This demonstrates the efficiency and scalability of our light weight feature vector design and offers promising chances for mobile and realtime usage.
Volatile classification rates in between different exercise types may be explained with the discrete value domain of our labels as well as with the subjective labeling procedure -- this assumption is also indicated by Figure \ref{fig:featuresandlunges}-2, which shows a confusion matrix for the qualitative assessment of bicycle crunches.
Incorrectly classified events became assigned to neighboring quality classes. 
Because an event's label originates from its rounded error score, it may occur that the label score is a border value, e.g. 2.5.
The event gets the label 3, although its quality is rated between 2 and 3. 
By contrast, the classifier may now decide that the feature vector looks more like a member of class 2, which finally leads to a wrong classification.

\subsection{Segmentation results}\label{subsec:segres}
All in all, we were able to extract 7,413 out of 7,534 recorded motion events into individual segments of specific length, using the segmentation approach described in Section \ref{subsec:segmentation}, which makes a total of 98,4\% of extracted events.  

\subsection{Activity recognition}\label{subsec:recognition}
Our preliminary study in \cite{ebert2017sensx} evaluated the automated recognition of eight different body weight exercises on basis of acceleration data. 
In this paper, our feature vectors are built on basis of the individual acceleration and rotation signals' standard deviations, rather than of time series with a fixed and significantly bigger length. 
We applied this new design to our dataset (see Section \ref{subsec:dataset}) and achieved a correct recognition rate of 99.9\% for manual classifier configuration (RF). 
100\% were reached for applying automated hyper parameter optimization within 10-fold cross validation -- training of the evaluation model took 4.9$s$ for 7,413 instances.  
Compared to our preliminary studies and related approaches, this performance can be regarded as state-of-art within the field of complex human motion recognition \cite{morris2014recofit, jiang2015human}.
%

\section{Conclusion and Future Work}\label{sec:conclusion}
In this paper we presented a generic approach for dynamic and individual segmentation of recurrent human motion events as well as for qualitative assessment of complex human motion.
For evaluation we recorded an exemplary dataset containing 7,534 repetitions of six different different body weight exercises and extracted them into multi-dimensional segments of individual length. 
Additionally, all segments were tagged with a quality label.
We are able to estimate a generic quality class of an individual event occurrence with an average correct classification rate of 99.3\% and up to 100\% for individual exercise types by adapting an expressive and heavily compressed feature vector.
For sheer recognition of activities, we actually reach a correct classification rate of 100\%. 
Automatic hyper parameter optimization performed significantly better than manual approaches.
Our concept features a generic analysis approach and we conjecture it is applicable to various recurrent human motions and transferable into multiple operational areas, such as sports, medical observation, or even workflow optimization.
\par
These results offer promising options for future work, e.g., a more fine-grained assessment process.
By adapting compressed feature vectors, information which is valuable for identifying tangible errors or the positioning of malpositions gets lost. 
Thereby, a conducted movement can be rated good or bad in a qualitative manner -- but neither can the exact reason for that be identified, nor can we carve out concrete characteristics of a specific quality assessment.  
New features and principal components may be crucial to explore these issues.
More dynamic and generic analysis approaches, e.g., neural networks, may bring new insights and are subject of ongoing research.

\bibliographystyle{IEEEtran}
\bibliography{refs}


\end{document}